\documentclass[12pt,nopreprintline]{elsarticle}

\usepackage{amssymb}
\usepackage{booktabs}
\usepackage{multirow}
\usepackage{xcolor}
\usepackage{soul}
\usepackage{url}

\journal{Artificial Intelligence in Medicine}

\begin{document}

\begin{frontmatter}

\title{Clinical trial cohort selection using Large Language Models on n2c2 Challenges}

\author[a]{Chi-en Amy Tai}
\affiliation[a]{organization={University of Waterloo, Vision and Image Processing Lab, VIP},
            addressline={200 University Avenue West},
            city={Waterloo},
            postcode={N2L3G1},
            country={Canada}}

\author[b]{Xavier Tannier}
\affiliation[b]{organization={Sorbonne Université, INSERM, Université Sorbonne Paris-Nord, Laboratoire d'Informatique Médicale et d'Ingénierie des Connaissances en eSanté, LIMICS},
            addressline={15 rue de l'école de médecine},
            city={Paris},
            postcode={75006},
            country={France}}

\begin{abstract}
Clinical trials are a critical process in the medical field for introducing new treatments and innovations. However, cohort selection for clinical trials is a time-consuming process that often requires manual review of patient text records for specific keywords. Though there have been studies on standardizing the information across the various platforms, Natural Language Processing (NLP) tools remain crucial for spotting eligibility criteria in textual reports. Recently, pre-trained large language models (LLMs) have gained popularity for various NLP tasks due to their ability to acquire a nuanced understanding of text. In this paper, we study the performance of large language models on clinical trial cohort selection and leverage the n2c2 challenges to benchmark their performance. Our results are promising with regard to the incorporation of LLMs for simple cohort selection tasks, but also highlight the difficulties encountered by these models as soon as fine-grained knowledge and reasoning are required.
\end{abstract}



\end{frontmatter}


\section{Introduction}
\label{introduction}
Clinical trials are a critical process in the medical field for introducing new treatments and innovations~\cite{weng2015optimizing}. However, cohort selection for clinical trials is a time-consuming process that often requires manual review of patient text records for specific keywords. Naturally, Clinical Trial Recruitment Support Systems have been developed as a large-scale platform that determines patient eligibility from federated Electronic Health Record (EHR) systems~\cite{doods2014piloting, soto2015efficiency, girardeau2017leveraging}. Unfortunately, EHR systems in different sites and different countries store patient information in different forms. Though there have been studies on standardizing the information across the various platforms~\cite{weng2010formal, luo2011dynamic, jiang2016developing, daniel2016cross}, these researchers mostly focused on rule-based definitions that struggle to incorporate the dynamic nature of clinical notes, which often contain details missing from the structured patient metadata~\cite{kirby2016phekb}. Another way of dealing with this issue, without standardizing the information, is to use natural language processing (NLP) techniques to match free text with inclusion and exclusion criteria. The n2c2 challenges catalyzed research focused on this idea, presenting datasets that contained patient clinical records and the associated correct values for various selection criteria for researchers to benchmark various approaches~\cite{n2c2datasets}. In these challenges, solutions mostly leveraged rule-based systems and some machine learning approaches~\cite{uzuner2008identifying,uzuner2009recognizing,stubbs2019cohort}.

Recently, pre-trained large language models (LLMs) have gained popularity for various NLP tasks due to their ability to acquire a nuanced understanding of text~\cite{eloundou2023gpts}. Trained on vast text datasets, these models appear to capture general understanding of concepts and with effective prompting, have demonstrated remarkable capabilities that could circumvent tedious manual tasks, such as concept extraction~\cite{kalyan2023survey}. Despite immense study on the performance of LLMs on various general and specific tasks, there has been some debate on their effectiveness when applied to specific tasks that may require specialized knowledge~\cite{karabacak2023embracing}.

In this paper, we study the performance of large language models on clinical trial cohort selection and leverage the n2c2 challenges to benchmark their performance. 
Section 2 presents more background about the n2c2 challenges and prompting techniques in research. The specific n2c2 datasets used in this study are detailed in Section 3 with the method described in Section 4. Section 5 presents the results for the model analysis and Section 6 discusses the findings from these results. A brief conclusion and ideas on future steps are written in Section 7.

\section{Related Work}
\label{related}
Since the recent arrival of LLMs, many articles have focused on their potential role in patient selection for clinical trials~\cite{Ghim2023TransformingCT}. Like us, most consider the objective of comparing patient information (structured or textual) with inclusion or exclusion criteria from available trials. LLMs can be used to convert free-text eligibility criteria to formal and structured elements, and then to use traditional NLP methods to match these elements~\cite{Wong2023ScalingCT}. They can also be used to convert eligibility criteria to ``EHR-like" sentences, \textit{i.e} sentences whose style corresponds to that found in medical documents~\cite{Yuan2023LargeLM}. One can finally ask the LLM to compare patients’ EHRs and eligibility critera, all at once~\cite{Nievas2023DistillingLL}. The datasets used, the methods compared and the evaluation criteria are very heterogeneous, and it is very difficult to really estimate the contribution of LLMs compared with more conventional methods. This evaluation problem is not specific to patient selection, but common to all areas of LLMs applied to the medical field~\cite{zaghir2024prompt}. That is why we specifically focused on n2c2 challenges, which have already been the subject of numerous studies in the past.

\subsection{n2c2 Challenges}
Challenges and competitions in the research community serve as catalysts for innovation, fostering collaboration, and driving advancements across various fields~\cite{cheng2020building}. In particular, the n2c2 challenges are an invaluable contribution in the medical natural language research community as they bring forth a challenge dataset that is of high quality and draw a multitude of academic institutions across the world to brainstorm innovative solutions for addressing pertinent healthcare problems~\cite{n2c2datasets}. 

The n2c2 challenge datasets represent a pivotal resource in the realm of clinical informatics, aimed at harnessing the potential of unstructured clinical text for advancing healthcare~\cite{n2c2datasets}. These datasets are designed to facilitate research that leverages existing clinical records to extract valuable insights, thereby fostering improvements in medical practice and patient outcomes. The interest in these datasets within the medical community stems from their ability to support natural language processing (NLP) and machine learning techniques tailored to healthcare contexts. By providing access to comprehensive, deidentified text from the Research Patient Data Registry at Partners, the n2c2 datasets enable researchers to tackle challenges such as disease recognition, information extraction, and temporal relations. Years after the competition has completed, these datasets are still being used to advance research and study the results of different approaches such as transformers~\cite{yang2020clinical}.

The challenge itself also harnesses great value as the corresponding competitors, their approaches, and their results are published in a summarized format for comparison at the end of each challenge~\cite{uzuner2008identifying,uzuner2009recognizing,stubbs2019cohort}. These summarized result papers are incredibly useful as they contain diverse solutions from competitors around the world along with a discussion of the reason why some solutions performed better than others. 

With relation to clinical trial cohort selection, there are three notable n2c2 challenge datasets that capture labeled records: 2006 - deidentification \& smoking~\cite{uzuner2008identifying}, 2008 - obesity~\cite{uzuner2009recognizing}, and 2018 (Track 1) - clinical trial cohort selection~\cite{stubbs2019cohort}. These challenge datasets have been used immensely in research to identify patient smoking status from medical discharge records, recognize obesity and co-morbidities in sparse data, and identify selection criteria for cohort selection. They are further discussed in the Datasets section. At the time of these competitions, submitted high-performing solutions mostly leveraged rule-based systems and some machine learning approaches. 

To the best of our knowledge, there have only been studies analyzing the performance of proprietary large language models such as GPT-3.5 and GPT-4 on the n2c2-2018 dataset with performance reportedly exceeding that previously stated in the original dataset papers~\cite{wornow2024zeroshotclinicaltrialpatient,Beattie2024}. The question of the legitimacy of these API-based approaches in relation to the conditions of use of this corpus is not discussed in these articles.

\subsection{Prompting Techniques}
Pre-trained large language models (LLMs) have recently gained popularity for natural language processing tasks, revolutionizing various fields from text generation to sentiment analysis and machine translation with a high labour market impact potential~\cite{eloundou2023gpts}. These models, such as OpenAI's GPT-3 and its predecessors, have demonstrated remarkable capabilities in understanding and generating human-like text for both academic research and industrial applications~\cite{brown2020language}. Trained on vast text datasets, these models acquire a nuanced understanding of language with adoption across a multitude of industries~\cite{kalyan2023survey}. However, when applied to more specialized cases, these generalized models inaccurately capture specific nuances in the field~\cite{karabacak2023embracing}. In the medical realm though, there have been releases of new models trained specifically on medical texts such as MedAlpaca~\cite{han2023medalpaca} to counteract the effects of training on generalized data. Though these pre-trained medical LLMs have shown potential, crafting effective prompts still remains an important design technique that greatly affects the LLMs' task performance~\cite{zamfirescu2023johnny}. 

The importance of prompts are stressed in  Beattie et al. and Wornow et al., both of whom analyzed the performance of GPT-3.5 and GPT-4 on the n2c2-2018 dataset~\cite{wornow2024zeroshotclinicaltrialpatient,Beattie2024}. In Beattie et al.~\cite{Beattie2024}, they applied a dynamic prompting strategy where they customize a generalized template using expert guidance for each criterion. They did not incorporate any direct prior examples in their prompt and instead, used a zero-shot learning approach to make classifications. Wornow et al.~\cite{wornow2024zeroshotclinicaltrialpatient} 
also designed a zero-shot system but investigated four different prompt strategies. Specifically, Wornow et al. tried the combinations of evaluating all the criteria or individual criteria with all the patient notes or individual patient notes. Comparing their results, Wornow et al. achieved the highest overall micro-F1 score using GPT-4 (0.93). Unfortunately Wornow et al. do not report their performance on each individual criteria and it is unclear if the reported GPT-4 results in Beattie et al. are for the entire test data.

As studied in Zaghir et al.~\cite{zaghir2024prompt}, the most recurrent prompt technique in medical applications is chain-of-thought (CoT). CoT leverages the prompt style ``Think step by step" to encourage the LLM to reason the answer. Instead of asking for a single answer, CoT encourages the LLM to split the problem into a series of intermediate steps and solve each one sequentially before producing the answer~\cite{wei2022chain}. When experimented on three large language models, CoT was shown to improve performance on a variety of tasks ranging from arithmetic to commonsense reasoning tasks~\cite{wei2022chain}. Interestingly, neither Beattie et al. nor Wornow et al. investigated CoT prompts in their analysis~\cite{Beattie2024,wornow2024zeroshotclinicaltrialpatient}.

In this study, we compare the performance of a basic straightforward prompt with few-shot learning incorporating CoT and then iterate on the few-shot learning examples to account for potential gaps and edge cases. 

\section{Datasets}
\label{data}
This section details the three challenge datasets used in our study: n2c2-2006, n2c2-2008, and n2c2-2018. 

\subsection{n2c2-2006 Challenge Dataset}
The n2c2-2006 challenge dataset is a publicly available dataset that contains randomly drawn de-identified clinical records from Partners HealthCare for identifying patient smoking status~\cite{uzuner2008identifying}. This dataset contained a total of 502 records with 398 in the training set and 104 in the test set. The records underwent tokenization, segmentation into sentences, conversion to XML format, and partitioning into training and test sets.  Two pulmonologists then annotated the data and classified patient records into one of five possible smoking status categories: past smoker (smoker one or more years ago but have not smoked for at least one year), current smoker (smoker within the past year), smoker (smoker but not enough information to classify as current or past), non-smoker (never smoked), and unknown. For classification, second-hand smokers were considered as non-smokers and activity around marijuana smoking was ignored as the labels only pertain with tobacco smoking. Given higher interannotor agreement performance for textual compared to intuitive annotations, it was decided that the records would be classified based only on explicitly mentioned information and any disagreements between the annotators were then resolved by obtaining judgments from two other pulmonologists. Records without a majority vote were then discarded from the dataset~\cite{uzuner2008identifying}. The distribution between the different patient categories are provided below for the training and test set in Table~\ref{tab:n2c2-2006-eda}. As seen, there are class imbalances in this dataset with some categories only having a few records (e.g., Smoker) while the Unknown category comprises more than half of the dataset. 

\begin{table}[!ht]
    \centering
    \begin{tabular}{lcc}
    \hline
    \textbf{Smoking Status} & \textbf{Training Data} & \textbf{Test Data} \\
    \hline
    Past Smoker & 36 & 11 \\
    Current Smoker & 35 & 11 \\
    Smoker & 9 & 3 \\
    Non-Smoker & 66 & 16 \\
    Unknown & 252 & 63 \\
    Total & 398 & 104 \\
    \hline
    \end{tabular}
    \caption{Distribution of the smoking categories in the n2c2-2006 challenge dataset~\cite{uzuner2008identifying}.}
    \label{tab:n2c2-2006-eda}
\end{table}

\subsection{n2c2-2008 Challenge Dataset}
The n2c2-2008 challenge dataset provides 1,237 discharge summaries that were de-identified semi-automatically with synthetic identifiers used to replace private health information~\cite{uzuner2009recognizing}. The patients in these summaries had been hospitalized for obesity or diabetes sometimes since 12/1/04 and were either overweight or diabetic from the Partners HealthCare Research Patient Data Repository. Fifteen obesity co-morbidities were identified: asthma, atherosclerotic cardiovascular disease (CAD), congestive heart failure (CHF), depression, diabetes mellitus (DM), gallstones/cholecystectomy, gastroesophageal reflux disease (GERD), gout, hypercholesterolemia, hypertension (HTN), hypertriglyceridemia, obstructive sleep apnea (OSA), osteoarthritis (OA), peripheral vascular disease (PVD), and venous insufficiency. Two obesity experts from the Massachusetts General Hospital Weight Center annotated the data and were given two types of tasks: textual and intuitive tasks. Textual tasks were to be labeled based on explicitly documented information in the summary and had four classifications: Present, Absent, Questionable, or Unmentioned. Intuitive tasks were based on intuition and judgment to information in the summary and had only three labels: Present, Absent, or Questionable. After annotation, a resident from the Massachusetts General Hospital resolved any disagreements in the textual judgments~\cite{uzuner2009recognizing}. 

\begin{figure}[!ht]
    \centering
    \includegraphics[width=\linewidth]{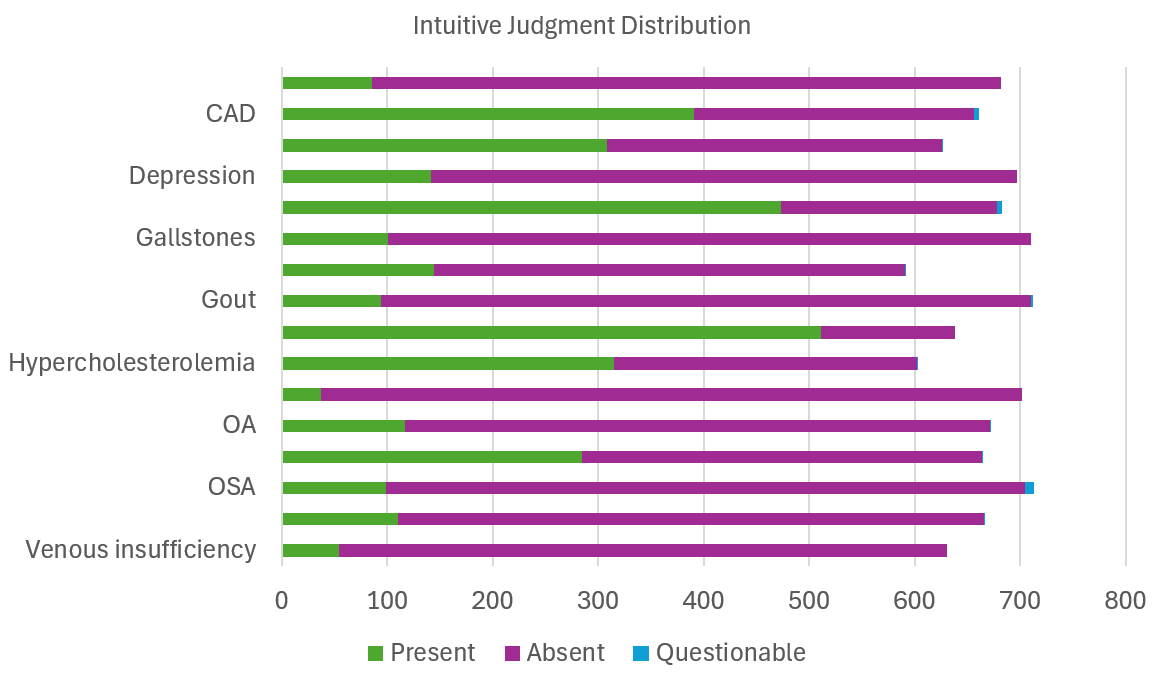}
    \caption{The n2c2-2008 dataset distribution for the fifteen intuitive criteria.}
    \label{fig:n2c2-2008-intuitive-data-distribution}
\end{figure}

The data contains non-uniform distributions for the classes in the training set is illustrated below in Figure~\ref{fig:n2c2-2008-textual-data-distribution} and Figure~\ref{fig:n2c2-2008-intuitive-data-distribution} for textual and intuitive tasks~\cite{uzuner2009recognizing}.

\begin{figure}[!ht]
    \centering
    \includegraphics[width=\linewidth]{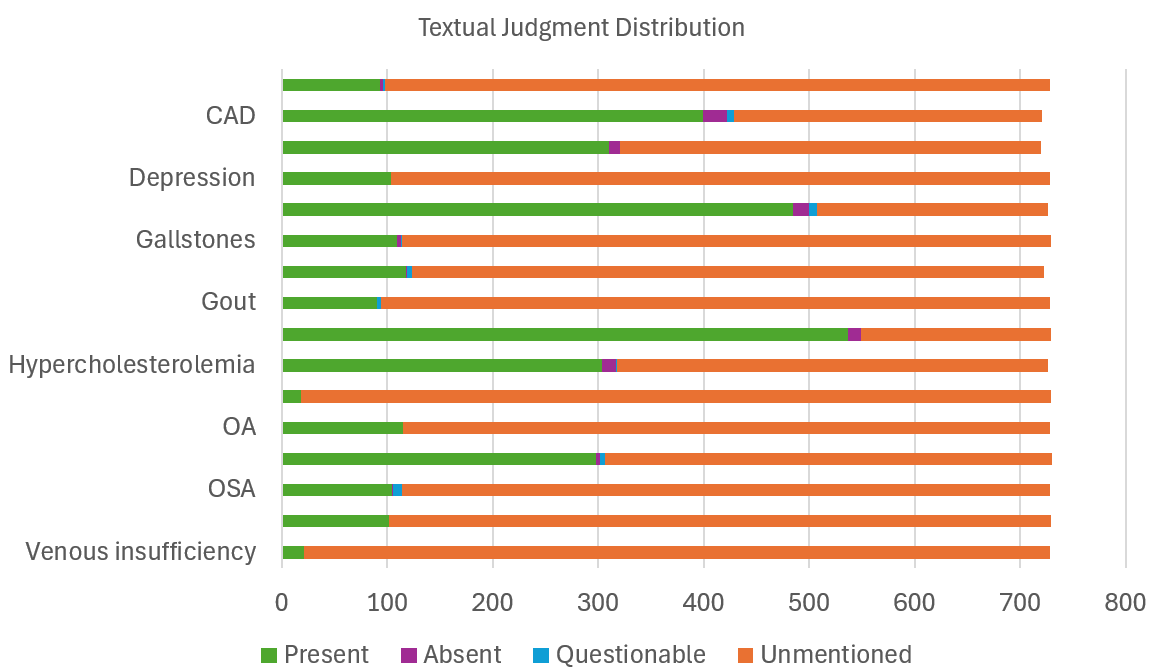}
    \caption{The n2c2-2008 dataset distribution for the fifteen textual criteria.}
    \label{fig:n2c2-2008-textual-data-distribution}
\end{figure}

\subsection{n2c2-2018 Challenge Dataset}
The n2c2-2018 challenge dataset contains 288 patients with 202 patients in the training set and 86 patients in the test set~\cite{stubbs2019cohort}. There are a total of 1,267 English records (781,006 tokens) with a rough estimate of 2 to 5 documents per patient (2,711 tokens per set of patient records). The records were obtained from a collection of American English longitudinal records, namely the 2014 i2b2/UTHealth shared tasks. All records were then de-identified as per the Health Insurance Portability Accountability Act guidelines with the removal and replacement of patient-linking information with realistic surrogates and random time-shifting of dates~\cite{stubbs2019cohort}. 

Since most of the patients in this dataset are at risk for heart disease and all of them have diabetes, the studied criteria were either common to most studies or related to these health conditions~\cite{stubbs2019cohort}. A total of thirteen criteria were selected: 
\begin{enumerate}
    \item ABDOMINAL: History of intra-abdominal surgery, small or large intestine resection, or small bowel obstruction
    \item ADVANCED-CAD: Advanced cardiovascular disease (CAD) defined by having two or more of the 4 sub-criteria defined in the guidelines
    \item ALCOHOL-ABUSE: Current alcohol use over weekly recommended limits
    \item ASP-FOR-MI: Use of aspirin to prevent MI
    \item CREATININE: Serum creatinine $>$ upper limit of normal
    \item DIETSUPP-2MOS: Taken a dietary supplement (excluding vitamin D) in the past 2 months
    \item DRUG-ABUSE: Drug abuse, current or past
    \item ENGLISH: Patient must speak English
    \item HBA1C: Any hemoglobin A1c (HbA1c) value between 6.5\% and 9.5\%
    \item KETO-1YR: Diagnosis of ketoacidosis in the past year
    \item MAJOR-DIABETES: Major diabetes-related complication - among 6 listed in the guidelines
    \item MAKES-DECISIONS: Patient must make their own medical decisions
    \item MI-6MOS: MI in the past 6 months
\end{enumerate}

\begin{figure}[!ht]
    \centering
    \includegraphics[width=\linewidth]{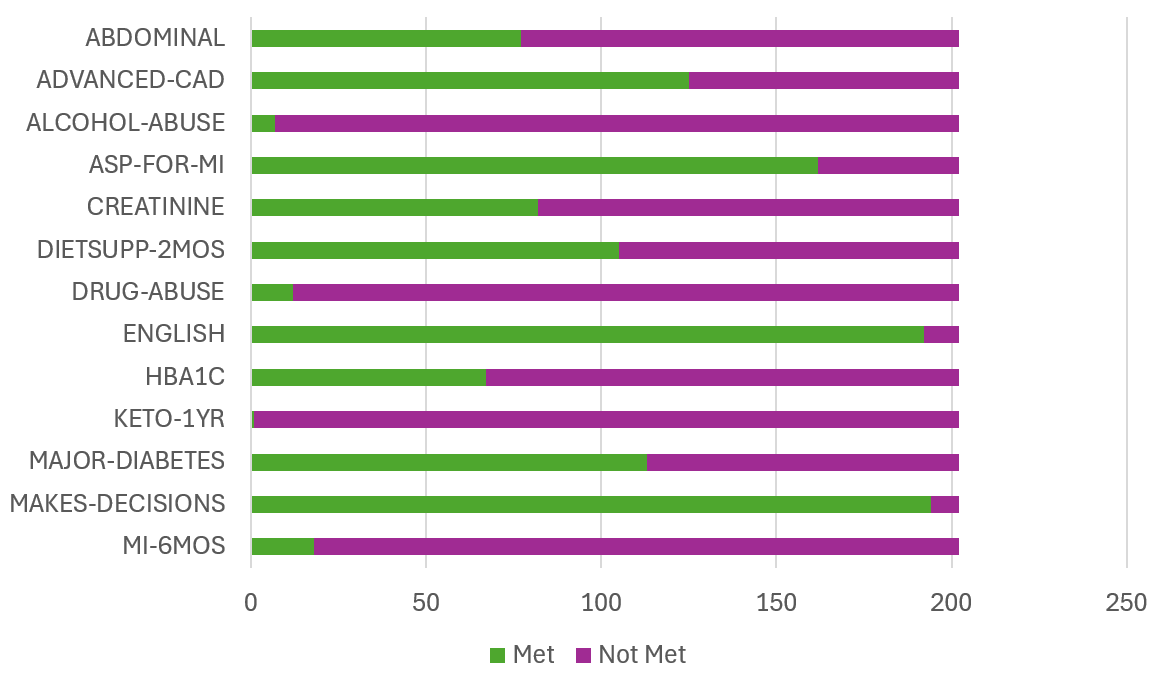}
    \caption{The n2c2-2018 dataset distribution for the thirteen criteria.}
    \label{fig:n2c2-2018-data-distribution}
\end{figure}

Two medical experts then independently annotated all the patient records and categorized each record as met or not met for thirteen criteria. The overall distribution for the thirteen criteria for the train set is shown in Figure~\ref{fig:n2c2-2018-data-distribution}. 

\section{Method}
\label{method}
A two-stage approach was implemented to evaluate and select the most effective large language model (LLM) for clinical trial cohort selection. As mentioned, three n2c2 challenge datasets were studied. In order to respect these datasets conditions of use, we  limited ourselves to LLMs that we could use on our own computing cluster.\footnote{``The data user will not disclose, disseminate, or otherwise share the Data / Datasets to or with any other
person or entity" \url{https://projects.iq.harvard.edu/files/n2c2/files/n2c2_data_sets_dua_preview_-_academic_user.pdf}}

In the first stage, various LLMs were systematically assessed based on their performance on a series of prompts for the n2c2-2018 dataset with the aim to identify the most promising LLM for clinical trial cohort selection. In the second stage, fine-tuning of the selected LLM from the first stage was conducted for the n2c2-2006 and n2c2-2008 challenge datasets.

\subsection{Stage 1: Selection of the Best LLM using n2c2-2018 Dataset}
The n2c2-2018 challenge dataset was utilized as the primary dataset for this stage. The original training set was divided into a new training set and a validation set to facilitate the model selection and tuning process. This split was conducted to ensure a robust evaluation of each model's performance in a controlled setting. 

Three different prompting strategies were tested across a range of LLMs to assess their capability in extracting relevant information for clinical trial cohort selection:
\begin{enumerate}
    \item \textbf{Initial Basic Prompt:} A straightforward prompt that outlined the task without any additional context or examples.
    \item \textbf{Few-Shot Learning:} A prompt that included a small number of example inputs and outputs from the training set to guide the model's responses.
    \item \textbf{Iterated Few-Shot Learning:} A refined version of few-shot learning, where the examples were iteratively selected from the training set based on previous performance to better train the model.
\end{enumerate}

Examples for the different prompting strategies for one of the 5-shot learning for the ASP-MI task are listed in the Supplementary Material A.

Each LLM was evaluated using the same three prompts on both the training and validation sets. The iterated few-shot learning approach consistently outperformed the other methods and hence, a five-fold cross-validation method was employed. Each fold contained an equal number of examples and had a distinct set of few-shot learning examples selected from the training set. Notably the ratio of positive to negative samples varied across folds due to the imbalanced distribution in the training set. Using five folds provided a more comprehensive assessment of the LLM's performance, ensuring that the selection of the optimal LLM was based on its ability to generalize across different data distributions. The performance of the LLMs were measured by the the average and standard deviation of their F1 score across the five folds on both the validation and test sets of the n2c2-2018 challenge dataset. The LLM with the best F1 performance was then selected for the next stage.

\subsection{Stage 2: Fine-tuning of the Selected Stage 1 LLM using n2c2-2006 and n2c2-2008 Datasets}
The n2c2-2006 and n2c2-2008 challenge datasets were utilized as the primary datasets for this stage. The original training set was divided into a new training set and a validation set for the fine-tuning process. An iterative approach was taken to select the best set of few-shot learning examples. This process involved evaluating the model's performance with various sets of examples selected from the training set and iterating based on the highest F1 score on the training and validation set. 

The best few-shot learning set was then selected based on the F1 performance on the training and validation set. The performance was then recorded for the test set for the n2c2-2006 and n2c2-2008 challenge dataset using the best few-shot learning set.

\section{Results}
\label{results}
The results reported below are for the model analysis on the n2c2-2018 challenge dataset for both the validation and test split for seven different LLM models (Table~\ref{tab:n2c2-2018-val} and Table~\ref{tab:n2c2-2018-test} respectively), and the n2c2-2006 and n2c2-2008 challenge test dataset for the vicuna-13b model (Table~\ref{tab:n2c2-2006-test}, Table~\ref{tab:n2c2-2008-textual-test}, and Table~\ref{tab:n2c2-2008-intuitive-test}).

\begin{table}[!ht]
\begin{tabular}{lcccc}
    \toprule
    \textbf{Model} & \textbf{ABDOM} & \textbf{MI-6MOS} & \textbf{DECISION} & \textbf{DIABETES} \\ 
    \midrule
    gpt-6j & 0.571 {\small(0.004)} & 0.140 {\small(0.000)} & 0.086 {\small(0.011)} & 0.750 {\small(0.004)} \\
    medalpaca-7b & 0.544 {\small(0.049)} & 0.146 {\small(0.014)} & 0.053 {\small(0.041)} & 0.667 {\small(0.116)} \\
    mistral-7b-instruct & 0.633 {\small(0.033)} & 0.387 {\small(0.122)} & 0.105 {\small(0.014)} & 0.803 {\small(0.012)} \\
    mistral-7b & 0.703 {\small(0.112)} & 0.293 {\small(0.120)} & 0.154 {\small(0.070)} & 0.807 {\small(0.038)} \\
    open-orca-mistral-7b & 0.500 {\small(0.189)} & 0.121 {\small(0.119)} & 0.000 {\small(0.000)} & 0.000 {\small(0.022)} \\
    vicuna-13b & 0.643 {\small(0.046)} & 0.546 {\small(0.101)} & 0.103 {\small(0.031)} & 0.804 {\small(0.013)} \\
    vicuna-7b & 0.609 {\small(0.050)} & 0.250 {\small(0.047)} & 0.191 {\small(0.056)} & 0.784 {\small(0.023)} \\
    \midrule
    \textbf{Model}& \textbf{HBA1C} & \textbf{ENGLISH} 
    & \textbf{DRUG} & \textbf{DIETSUPP} \\
    \midrule
    gpt-6j & 0.788 {\small(0.074)} & 0.132 {\small(0.026)} & 0.080 {\small(0.014)} & 0.720 {\small(0.022)} \\
    medalpaca-7b & 0.596 {\small(0.221)} & 0.000 {\small(0.078)} & 0.083 {\small(0.099)} & 0.636 {\small(0.185)} \\
    mistral-7b-instruct & 0.348 {\small(0.216)} & 0.571 {\small(0.090)} & 0.462 {\small(0.100)} & 0.833 {\small(0.194)} \\
    mistral-7b & 0.447 {\small(0.204)} & 0.556 {\small(0.190)} & 0.353 {\small(0.194)} & 0.784 {\small(0.213)} \\
    open-orca-mistral-7b & 0.254 {\small(0.222)} & 0.143 {\small(0.156)} & 0.200 {\small(0.234)} & 0.241 {\small(0.230)} \\
    vicuna-13b & 0.467 {\small(0.263)} & 0.714 {\small(0.082)} & 0.600 {\small(0.146)} & 0.750 {\small(0.104)} \\
    vicuna-7b & 0.453 {\small(0.150)} & 0.385 {\small(0.152)} & 0.333 {\small(0.074)} & 0.761 {\small(0.056)} \\
    \midrule
    \textbf{Model} & \textbf{CREAT} & \textbf{ASP-MI} & \textbf{ALCOHOL} & \textbf{ADV-CAD}\\
    \midrule
    gpt-6j & 0.613 {\small(0.042)} & 0.873 {\small(0.136)} & 0.080 {\small(0.023)} & 0.750 {\small(0.094)} \\
    medalpaca-7b & 0.423 {\small(0.130)} & 0.606 {\small(0.244)} & 0.133 {\small(0.076)} & 0.756 {\small(0.008)} \\
    mistral-7b-instruct & 0.640 {\small(0.099)} & 0.827 {\small(0.030)} & 0.400 {\small(0.137)} & 0.772 {\small(0.039)} \\
    mistral-7b & 0.667 {\small(0.173)} & 0.596 {\small(0.249)} & 0.333 {\small(0.084)} & 0.743 {\small(0.080)} \\
    open-orca-mistral-7b & 0.238 {\small(0.094)} & 0.162 {\small(0.167)} & 0.182 {\small(0.250)} & 0.000 {\small(0.000)} \\
    vicuna-13b & 0.698 {\small(0.065)} & 0.880 {\small(0.034)} & 0.333 {\small(0.065)} & 0.747 {\small(0.031)} \\
    vicuna-7b & 0.615 {\small(0.072)} & 0.886 {\small(0.049)} & 0.286 {\small(0.106)} & 0.729 {\small(0.050)} \\
    \hline
\end{tabular}
\caption{5-shot results for the n2c2-2018 tasks: for each task and model, median and standard deviation over 5 runs are provided on the val set.}
\label{tab:n2c2-2018-val}
\end{table}

\begin{table}[!ht]
\begin{tabular}{lcccc}
    \toprule
    \textbf{Model} & \textbf{ABDOM} & \textbf{MI-6MOS} & \textbf{DECISION} & \textbf{DIABETES} \\ 
    \midrule
    gpt-6j & 0.527 {\small(0.027)} & 0.170 {\small(0.000)} & 0.072 {\small(0.012)} & 0.667 {\small(0.023)} \\
    medalpaca-7b & 0.457 {\small(0.053)} & 0.191 {\small(0.014)} & 0.066 {\small(0.047)} & 0.694 {\small(0.100)} \\
    mistral-7b-instruct & \textbf{0.602 {\small(0.032)}} & 0.356 {\small(0.079)} & 0.091 {\small(0.033)} & 0.746 {\small(0.041)} \\
    mistral-7b & 0.580 {\small(0.035)} & 0.372 {\small(0.134)} & \textbf{0.118 {\small(0.028)}} & 0.717 {\small(0.036)} \\
    open-orca-mistral-7b & 0.500 {\small(0.161)} & 0.222 {\small(0.084)} & 0.000 {\small(0.000)} & 0.000 {\small(0.020)} \\
    vicuna-13b & 0.556 {\small(0.064)} & \textbf{0.414 {\small(0.069)}} & 0.083 {\small(0.023)} & \textbf{0.791 {\small(0.031)}} \\
    vicuna-7b & 0.542 {\small(0.031)} & 0.259 {\small(0.055)} & 0.074 {\small(0.046)} & 0.688 {\small(0.025)} \\
    \midrule
    best from campaign & \textbf{0.906} & \textbf{0.876} & \textbf{0.897} & \textbf{0.884}  \\
    \midrule
    \textbf{Model}& \textbf{HBA1C} & \textbf{ENGLISH} 
    & \textbf{DRUG} & \textbf{DIETSUPP} \\
    \midrule
    gpt-6j & \textbf{0.735 {\small(0.091)}} & 0.265 {\small(0.030)} & 0.069 {\small(0.014)} & 0.650 {\small(0.030)} \\
    medalpaca-7b & 0.568 {\small(0.148)} & 0.000 {\small(0.092)} & 0.076 {\small(0.043)} & 0.631 {\small(0.137)} \\
    mistral-7b-instruct & 0.411 {\small(0.210)} & 0.667 {\small(0.066)} & \textbf{0.235 {\small(0.146)}} & 0.659 {\small(0.165)} \\
    mistral-7b & 0.494 {\small(0.150)} & 0.750 {\small(0.218)} & 0.100 {\small(0.096)} & 0.710 {\small(0.160)} \\
    open-orca-mistral-7b & 0.324 {\small(0.208)} & 0.400 {\small(0.193)} & 0.000 {\small(0.029)} & 0.087 {\small(0.224)} \\
    vicuna-13b & 0.405 {\small(0.192)} & \textbf{0.765 {\small(0.056)}} & 0.222 {\small(0.103)} & \textbf{0.730 {\small(0.124)}} \\
    vicuna-7b & 0.535 {\small(0.133)} & 0.615 {\small(0.102)} & 0.143 {\small(0.101)} & 0.667 {\small(0.056)} \\
    \midrule
    best from campaign & \textbf{0.950} & \textbf{0.977} & \textbf{0.920} & \textbf{0.919}  \\
    \midrule
    \textbf{Model} & \textbf{CREAT} & \textbf{ASP-MI} & \textbf{ALCOHOL} & \textbf{ADV-CAD}\\
    \midrule
    gpt-6j & 0.427 {\small(0.048)} & 0.818 {\small(0.112)} & 0.100 {\small(0.021)} & 0.687 {\small(0.033)} \\
    medalpaca-7b & 0.418 {\small(0.070)} & 0.569 {\small(0.262)} & 0.133 {\small(0.091)} & 0.683 {\small(0.007)} \\
    mistral-7b-instruct & \textbf{0.622 {\small(0.109)}} & \textbf{0.875 {\small(0.023)}} & 0.286 {\small(0.140)} & \textbf{0.719 {\small(0.022)}} \\
    mistral-7b & 0.512 {\small(0.142)} & 0.673 {\small(0.172)} & 0.286 {\small(0.140)} & 0.667 {\small(0.081)} \\
    open-orca-mistral-7b & 0.188 {\small(0.147)} & 0.225 {\small(0.118)} & 0.167 {\small(0.360)} & 0.000 {\small(0.000)} \\
    vicuna-13b & 0.605 {\small(0.060)} & 0.849 {\small(0.030)} & \textbf{0.375 {\small(0.103)}} & 0.716 {\small(0.030)} \\
    vicuna-7b & 0.581 {\small(0.061)} & \textbf{0.878 {\small(0.034)}} & 0.261 {\small(0.082)} & 0.644 {\small(0.054)} \\
    \midrule
    best from campaign & \textbf{0.898} & 0.770 & \textbf{0.897} & \textbf{0.870}  \\
    \hline
\end{tabular}
\caption{5-shot results for the n2c2-2018 tasks: for each task and model, median and standard deviation over 5 runs are provided on the test set. The best result from the official campaign is also provided.  The top two scores for each task are also bolded.}
\label{tab:n2c2-2018-test}
\end{table}

\begin{table}[!ht]
    \centering
    \begin{tabular}{lcc}
    \hline
    \textbf{Model} & \textbf{Macro F1} & \textbf{Micro F1} \\
    \hline
    vicuna-13b & 0.703 & 0.865 \\
    mistral-7b-instruct & 0.703 & 0.827 \\
    best from campaign & 0.760 & 0.900 \\
    \hline
    \end{tabular}
    \caption{Model performance for vicuna-13b and mistral-7b-instruct on n2c2-2006 task to categorize all smoker classes compared to the best result from the official campaign on the test set.}
    \label{tab:n2c2-2006-test}
\end{table}

\begin{table}[!ht]
    \centering
    \begin{tabular}{lccccccc}
    \hline
    \multirow{2}{*}{\textbf{Textual Task}} & \multicolumn{3}{c}{\textbf{Macro F1}} & \multicolumn{3}{c}{\textbf{Micro F1}} \\
     & \textbf{V13b}  & \textbf{M7b} & \textbf{Best} & \textbf{V13b} & \textbf{M7b} & \textbf{Best} \\
    \hline
    Asthma & 0.545 & 0.570 & 0.943 & 0.873 & 0.903 & 0.992 \\
    CAD & 0.534 & 0.460 & 0.856 & 0.799 & 0.738 & 0.926 \\
    CHF & 0.520 & 0.612 & 0.794 & 0.903 & 0.790 & 0.936 \\
    Depression & 0.459 & 0.472 & 0.972 & 0.953 & 0.964 & 0.984 \\
    Diabetes & 0.751 & 0.727 & 0.903 & 0.921 & 0.934 & 0.976 \\
    Gallstones & 0.561 & 0.331 & 0.814 & 0.921 & 0.613 & 0.982 \\
    GERD & 0.431 & 0.328 & 0.488 & 0.873 & 0.694 & 0.988 \\
    Gout & 0.439 & 0.435 & 0.973 & 0.927 & 0.933 & 0.988 \\
    Hypercholesterolemia & 0.518 & 0.566 & 0.792 & 0.841 & 0.779 & 0.972 \\
    Hypertension & 0.620 & 0.580 & 0.838 & 0.912 & 0.864 & 0.962 \\
    Hypertriglyceridemia & 0.297 & 0.546 & 0.973 & 0.890 & 0.850 & 0.998 \\
    OA & 0.544 & 0.558 & 0.959 & 0.875 & 0.892 & 0.976 \\
    Obesity & 0.426 & 0.457 & 0.488 & 0.809 & 0.892 & 0.968 \\
    OSA & 0.676 & 0.675 & 0.878 & 0.954 & 0.905 & 0.992 \\
    PVD & 0.555 & 0.434 & 0.968 & 0.888 & 0.734 & 0.986 \\
    Venous Insufficiency & 0.298 & 0.408 & 0.840 & 0.872 & 0.972 & 0.982 \\
    \hline
    \end{tabular}
    \caption{Model performance on n2c2-2008 textual tasks using vicuna-13b (V13b) and mistral-7b-instruct (M7b) on the test set.}
    \label{tab:n2c2-2008-textual-test}
\end{table}

\begin{table}[!ht]
    \centering
    \begin{tabular}{lccccccc}
    \hline
    \multirow{2}{*}{\textbf{Intuitive Task}} & \multicolumn{3}{c}{\textbf{Macro F1}} & \multicolumn{3}{c}{\textbf{Micro F1}} \\
     & \textbf{V13b}  & \textbf{M7b} & \textbf{Best} & \textbf{V13b} & \textbf{M7b} & \textbf{Best} \\
    \hline
    Asthma & 0.579 & 0.823 & 0.978 & 0.866 & 0.887 & 0.989 \\
    CAD & 0.603 & 0.559 & 0.612 & 0.902 & 0.852 & 0.919 \\
    CHF & 0.584 & 0.574 & 0.624 & 0.836 & 0.856 & 0.932 \\
    Depression & 0.536 & 0.736 & 0.935 & 0.839 & 0.790 & 0.954 \\
    Diabetes & 0.628 & 0.634 & 0.968 & 0.912 & 0.958 & 0.973 \\
    Gallstones & 0.633 & 0.523 & 0.973 & 0.963 & 0.839 & 0.986 \\
    GERD & 0.462 & 0.477 & 0.577 & 0.622 & 0.683 & 0.913 \\
    Gout & 0.556 & 0.587 & 0.977 & 0.844 & 0.914 & 0.990 \\
    Hypercholesterolemia & 0.553 & 0.576 & 0.905 & 0.784 & 0.863 & 0.907 \\
    Hypertension & 0.586 & 0.897 & 0.885 & 0.919 & 0.937 & 0.928 \\
    Hypertriglyceridemia & 0.365 & 0.388 & 0.798 & 0.774 & 0.817 & 0.971 \\
    OA & 0.556 & 0.544 & 0.629 & 0.799 & 0.866 & 0.959 \\
    Obesity & 0.589 & 0.609 & 0.972 & 0.852 & 0.913 & 0.973 \\
    OSA & 0.627 & 0.621 & 0.881 & 0.925 & 0.960 & 0.994 \\
    PVD & 0.500 & 0.575 & 0.635 & 0.746 & 0.886 & 0.976 \\
    Venous Insufficiency & 0.417 & 0.427 & 0.808 & 0.735 & 0.855 & 0.962 \\
    \hline
    \end{tabular}
    \caption{Model performance on n2c2-2008 intuitive tasks using vicuna-13b (V13b) and mistral-7b-instruct (M7b) on the test set.}
    \label{tab:n2c2-2008-intuitive-test}
\end{table}

\section{Discussion}
\label{discussion}

\subsection{Lessons learned}
The results showcased that using LLMs for patient cohort analysis are promising, with some LLMs generally performing better than others for cohort selection. Compared to the reported performance on the n2c2-2006, n2c2-2008, and n2c2-2018 datasets, the best LLM was able to achieve better or similar performance for some of the tasks. For example, for the n2c2-2008 intuitive hypertension task, mistral-7b-instruct achieved a better performance than that reported on the n2c2-2008 dataset paper. On the other hand, for others like the n2c2-2008 intuitive venous insufficiency, the best LLM achieved a lower performance compared to those reported from the n2c2 challenge. Interestingly, both mistral-7b-instruct and vicuna-13b perform well on the n2c2-2018 tasks, though vicuna-13b often has similar results with more stability (lower standard deviation) across the tasks. Notably, one should be careful when employing LLMs for generalized patient cohort selection tasks, especially for scenarios with high imbalance and potential nuances such as identifying if the patient has abdominal issues or makes decisions. However, for selection criteria that are fairly straightforward and/or have regular examples such as classifying the patient's smoking class or if they use aspirin for MI, using an LLM could be very useful for cohort selection. 

These mixed results are consistent with other studies on clinical information retrieval with LLMs, showing that fine-grained knowledge and reasoning mechanisms are not yet sufficient for real-life use cases \cite{nagar2024llmszeroshotreasonersbiomedical,naguib2024shotclinicalentityrecognition}.

Overall, in this rigorous evaluation and comparison with more traditional techniques, our results run counter to a common view that generative LLMs generally outperform existing methods, or are even likely to solve the problem of patient selection once and for all.

\subsection{Limitations}
This work explores the classic and general uses of LLMs, but has certain limitations.
In particular, we have only applied approaches that do not require any specific adaptation to each variable researched. Chains of thoughts, prompt optimization~\cite{opsahlong2024optimizinginstructionsdemonstrationsmultistage,soylu2024finetuningpromptoptimizationgreat} or other techniques could certainly lead to better results, but would be difficult to apply in a generic selection process.

As already mentioned, we did not use commercial LLMs (such as ChatGPT or Gemini), as their standard conditions of use are not compatible with working on confidential medical reports.

\section{Conclusion}
\label{conclusion}
This paper studied the performance of large language models on clinical trial cohort selection and benchmarked various models on three n2c2 challenge datasets against previous non-LLM approaches. Though there are promising results for some of the n2c2 challenge tasks, it is evident that LLMs perform better when the selection criteria is straightforward and poorly when there are more nuances with the selection criteria. Surprisingly, the results also indicate that vicuna-13b and mistral-7b-instruct models (trained on general texts) perform the best with more stability across the n2c2-2018 cohort selection tasks compared to the other studied LLM models such as MedAlpaca (trained on medical texts). 

Extensions of this work thus include generalizing the process of cohort selection using LLM across a variety of domains, such as that discussed in~\cite{datta2024autocriteria}; and fine-tuning an LLM using clinical text records, similar to that in ClinicalMamba~\cite{yang2024clinicalmamba}. 

\bibliographystyle{ieeetr}
\bibliography{refs}

\section*{Supplementary Material A: Prompt examples}

\{\} is used to denote the location of the selected patient text for inference.

\noindent\makebox[\linewidth]{\rule{\linewidth}{0.4pt}}
\noindent \textbf{Initial Basic Prompt}

\noindent 
\color{blue}
\textit{These are some sentences from a patient's clinical report. Answer Yes or No to the final question. \\ \\
\{\}
\\ \\
Question: Does the text mention that the patient uses aspirin to prevent myocardial infarction (MI)?}

\color{black}

\noindent\makebox[\linewidth]{\rule{\linewidth}{0.4pt}}
\noindent \textbf{Few-Shot Learning}

\noindent 
\color{blue}
\textit{These are some sentences from a patient's clinical report. Does the text mention that the patient uses aspirin to prevent myocardial infarction (MI)? Answer Yes or No to the final question. Let's think step by step.\\ \\
Context: "MEDICATIONS:  Aspirin 81 mg p.o. daily, calcium carbonate 600 mg b.i.d., Diovan 80 mg daily, Glucophage 850 mg b.i.d., lorazepam 1 mg q.i.d. p.r.n., Paxil 10 mg daily, and fluvastatin 20 mg daily, but she ran out some time ago."\\
Answer: The text mentions a prescription of aspirin with a dose of 81 mg daily. 81 mg is a low dose (less than 325 mg). So the answer is Yes.\\
Context: "MEDICATIONS: Synthroid, Hydralazine, Lopressor, prednisone, Coumadin,"\\
Answer: The text mentions several medications, but does not mention aspirin. So the answer is No.\\
Context: "EMS gave her 3 puffs of NTG and 4 baby ASA w/o effect. "\\
Answer: The text mentions a prescription of baby aspirin (ASA). Baby aspirin is a common name for low-dose aspirin. So the answer is Yes.\\
Context: "1. Aspirin."\\
Answer: The text mentions a prescription of aspirin, but does not mention the dose. So the answer is No.\\
Context: "Ecotrin 325 mg PO QD"\\
Answer: The text mentions a prescription of aspirin (Ecotrin) with a dose of 325 mg daily (QD). 325 mg is a low dose (less than 325 mg). So the answer is Yes.\\
Context: "ALLERGIES:  She has an allergy to codeine, aspirin, erythromycin,"\\
Answer: The text mentions an allergy to aspirin. So the answer is No.
Context: "he remains on a plethora of medications including NPH insulin 64 units sub q qAM along with 34 units CZI sub q qAM and 68 units NPH sub q qPM, enteric coated aspirin one tab po q.d., Captopril 25 mg po t.i.d., Lasix 40 mg po b.i.d.",\\
Answer: The text mentions a prescription of aspirin with a dose of one tab daily. This is not a low dose. So the answer is No. \\ \\
\{\}}
\color{black}

\noindent\makebox[\linewidth]{\rule{\linewidth}{0.4pt}}
\noindent \textbf{Iterated Few-Shot Learning}

\noindent 
\color{blue}
\textit{These are some sentences from a patient's clinical report. Does the text mention that the patient uses aspirin to prevent myocardial infarction (MI)? Answer Yes or No to the final question. Let's think step by step. \\ \\
Context: "MEDICATIONS:  Aspirin 81 mg p.o. daily, calcium carbonate 600 mg b.i.d., Diovan 80 mg daily, Glucophage 850 mg b.i.d., lorazepam 1 mg q.i.d. p.r.n., Paxil 10 mg daily, and fluvastatin 20 mg daily, but she ran out some time ago."\\
Answer: The text mentions a prescription of aspirin with a dose of 81 mg daily. 81 mg is a low dose (less than 325 mg). So the answer is Yes.\\
Context: "MEDICATIONS: Synthroid, Hydralazine, Lopressor, prednisone, Coumadin,"\\
Answer: The text mentions several medications, but does not mention aspirin. So the answer is No.\\
Context: "EMS gave her 3 puffs of NTG and 4 baby ASA w/o effect. "\\
Answer: The text mentions a prescription of baby aspirin (ASA). Baby aspirin is a common name for low-dose aspirin. So the answer is Yes.\\
Context: "1. Aspirin."\\
Answer: The text mentions a prescription of aspirin, but does not mention the dose. So the answer is No.\\
Context: "Ecotrin 325 mg PO QD"\\
Answer: The text mentions a prescription of aspirin (Ecotrin) with a dose of 325 mg daily (QD). 325 mg is a low dose (less than 325 mg). So the answer is Yes.\\
Context: "ALLERGIES:  She has an allergy to codeine, aspirin, erythromycin,"\\
Answer: The text mentions an allergy to aspirin. So the answer is No.\\ \\
\{\}}
\color{black}

\noindent\makebox[\linewidth]{\rule{\linewidth}{0.4pt}}

\end{document}